\documentclass[US letter, 10pt, conference]{ieeeconf}      

\IEEEoverridecommandlockouts

\usepackage{amsmath,amsfonts,amssymb} 
\usepackage{algorithmic}
\usepackage{algorithm}
\usepackage{array}
\usepackage[caption=false,font=normalsize,labelfont=sf,textfont=sf]{subfig}
\usepackage{textcomp}
\usepackage{stfloats}
\usepackage{url}
\usepackage{verbatim}
\usepackage{graphicx}
\usepackage{cite}
\def\BibTeX{{\rm B\kern-.05em{\sc i\kern-.025em b}\kern-.08em
    T\kern-.1667em\lower.7ex\hbox{E}\kern-.125emX}}
\usepackage[pdfstartview=XYZ,
bookmarks=true,
colorlinks=true,
linkcolor=black,
urlcolor=black,
citecolor=black,
bookmarks=true,
linktocpage=true, 
hyperindex=true
]{hyperref}
\usepackage{balance}

\usepackage{orcidlink}
   
\begin{document}

\title{\LARGE \bf
Immersive Teleoperation of Beyond-Human-Scale Robotic Manipulators: Challenges and Future Directions
}

\author{Mahdi Hejrati \orcidlink{0000-0002-8017-4355} and Jouni Mattila \orcidlink{0000-0003-1799-4323}
\thanks{This work is supported by Business Finland partnership project "Future all-electric rough terrain autonomous mobile manipulators" (Grant 2334/31/222). Corresponding author: Mahdi Hejrati}
\thanks{All authors are with the Department of Engineering and Natural Science, Tampere University, 7320 Tampere, Finland (e-mail: mahdi.hejrati@tuni.fi, jouni.mattila@tuni.fi).} %
}

© 2025 IEEE. Personal use of this material is permitted.
Permission from IEEE must be obtained for all other uses,
including reprinting/republishing this material for advertising
or promotional purposes, collecting new collected works
for resale or redistribution to servers or lists, or reuse of
any copyrighted component of this work in other works.
This work has been submitted to the IEEE for possible
publication. Copyright may be transferred without notice,
after which this version may no longer be accessible.

This work has been accepted for presentation at the 2025 IEEE Conference on Telepresence, to be held in Leiden, Netherlands.
\maketitle
\thispagestyle{empty}
\pagestyle{empty}

\begin{abstract}
Teleoperation of beyond-human-scale robotic manipulators (BHSRMs) presents unique challenges that differ fundamentally from conventional human-scale systems. As these platforms gain relevance in industrial domains such as construction, mining, and disaster response, immersive interfaces must be rethought to support scalable, safe, and effective human–robot collaboration. This paper investigates the control, cognitive, and interface-level challenges of immersive teleoperation in BHSRMs, with a focus on ensuring operator safety, minimizing sensorimotor mismatch, and enhancing the sense of embodiment. We analyze design trade-offs in haptic and visual feedback systems, supported by early experimental comparisons of exoskeleton- and joystick-based control setups. Finally, we outline key research directions for developing new evaluation tools, scaling strategies, and human-centered safety models tailored to large-scale robotic telepresence.
\end{abstract}

\section{INTRODUCTION}

With the rapid and successful advancement of machine learning algorithms in real-world applications, a critical challenge remains: \textit{how to collect high-quality, representative data from diverse robotic systems to enable effective learning and deployment}. As the human operator is typically the sole expert demonstrator in teleoperation settings, the primary goal of machine learning-based control is to enable the transfer of human-level skills to robotic systems, and eventually extend performance beyond human capabilities through policy optimization techniques, such as inverse reinforcement learning. Achieving such a transfer requires that the machine learning algorithms be trained on rich, high-quality datasets that accurately capture the nuances of human operation and decision-making through comprehensive multimodal information \cite{chi2023diffusion}. 

Since the effectiveness of policy training depends not only on the quantity of data but critically on its quality, it is essential to design teleoperation platforms that enable the human operator to act in a natural and realistic manner---that is, in a human-like way. Slater~\cite{slater2009place} argued that such realistic behavior in virtual environments emerges from the co-occurrence of two perceptual phenomena: \textit{place illusion}---the feeling of being physically present in the virtual space---and \textit{plausibility illusion}---the sense that the virtual events are actually occurring. Building on this framework, ~\cite{kilteni2012sense} introduced the concept of the \textit{sense of embodiment} (SoE), which describes the subjective experience of owning and controlling a remote body as if it were part of one's own. This sense is considered a critical factor in eliciting natural motor behaviors during teleoperation. The SoE comprises three core components: (1) \textit{body ownership}—the perception that the surrogate system is part of one's own body; (2) \textit{self-location}—the experience of being spatially present at the remote site; and (3) \textit{agency}—the sense that one's actions directly cause movements or effects in the environment. Among these, \textit{agency} and \textit{self-location} are particularly critical in real-world teleoperation scenarios, especially in industrial or high-risk field applications, as they directly influence task fluency, control precision, and overall operator performance~\cite{falcone2023toward}.

One effective way to elicit a strong SoE during teleoperation is to ensure a high degree of immersion within the control environment. As discussed by Slater~\cite{slater2009place}, a teleoperation system is considered more immersive than another if it provides superior sensory fidelity---such as higher visual display resolution, reduced latency, or enhanced sensorimotor congruency. Increased immersion fosters a stronger sense of presence and body ownership, thereby promoting more natural and realistic operator behaviors. This, in turn, enables the collection of high-quality, multimodal demonstration data that better reflects human intent and motor adaptation, thereby improving the training and generalization of learning algorithms.

While numerous immersive teleoperation frameworks have been developed to support learning from demonstration (LfD) in human-scale robotic systems~\cite{cheng2024open, iyer2024open}, comparatively less attention has been devoted to beyond-human-scale robotic manipulators (BHSRMs). These systems—including heavy-duty manipulators used in mining, construction, and agriculture—pose unique operational and cognitive challenges due to their size, power, and dissimilarity from the human body, which make intuitive control and safe human interaction far more difficult than in conventional settings. This is particularly evident in domains like mining, where teleoperation has been employed for large excavator machines used in drilling and tunneling, as well as for exploration and rescue robots deployed in hazardous or confined environments~\cite{lichiardopol2007survey}. The relevance of such systems is underscored by the scale of their economic impact: in 2023 alone, the combined revenue of the world’s top 40 mining companies reached 845 billion USD, with demand for mined materials projected to continue rising~\cite{hasan2025wireless}.

Although recent work has addressed aspects of high-performance bilateral control and shared autonomy for BHSRMs teleoperation~\cite{suomalainen2018learning, luo2022human, cheng2024multitarget, lampinen2021force, liu2023multi}, and several reviews have surveyed teleoperation strategies for excavators and mining machinery~\cite{hasan2025wireless, lee2022challenges}, critical questions remain regarding the role of human immersion and cognitive integration in these systems. Given that these large-scale machines constitute a significant portion of the industrial robotics landscape, realizing their automation potential hinges on the ability to elicit natural, high-quality demonstrations from human operators—particularly in unstructured and safety-critical environments. This requires teleoperation interfaces that are immersive enough to foster realistic behavior and evoke a strong SoE, thereby ensuring the collection of rich, high-fidelity multimodal data for policy learning. For this end, this paper aims to outline and critically examine the core control, perceptual, and human-centered challenges in achieving immersive teleoperation for BHSRMs.

\section{Control Challenges in Teleoperation of Beyond-Human-Scale Robots} 
In this section, we identify and analyze key control-related challenges that arise in the teleoperation of BHSRMs. Addressing these issues is essential for developing scalable, safe, and effective teleoperation frameworks that can support future automation in industrial and unstructured environments.

\begin{figure}
    \centering
    \includegraphics[width=\linewidth]{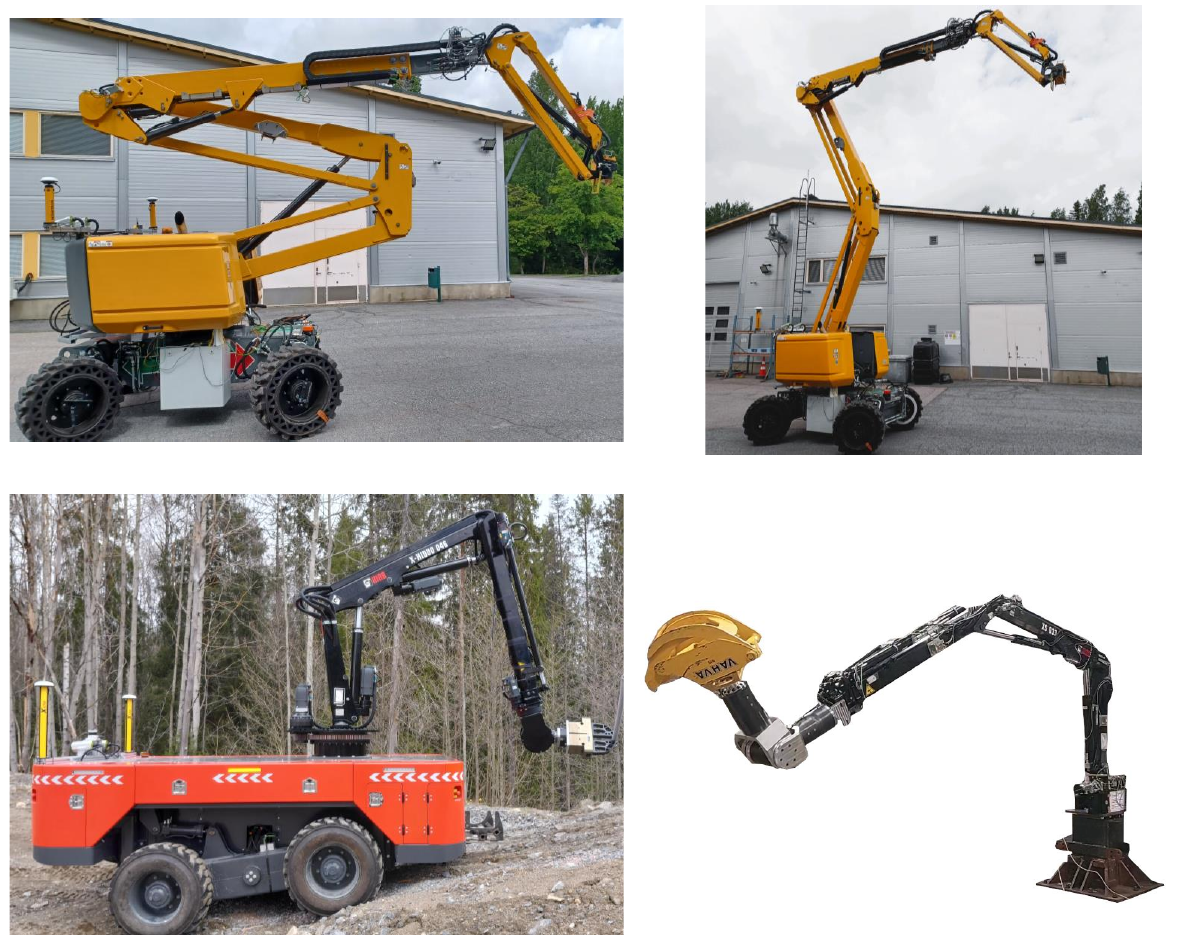}
    \caption{ Beyond-human-scale robotic manipulator operating in an unstructured terrain. These systems introduce unique control challenges such as high inertia, hydraulic nonlinearities, slippage, and delay-sensitive remote operation.}
    \label{fig:HHM}
\end{figure}

\subsection{Robustness and Transparency}
BHSRMs pose considerable control challenges due to their substantial inertia, nonlinear dynamics, hazardous operating conditions, structural complexity, friction, and unmodeled uncertainties. These challenges are further exacerbated in hydraulically actuated manipulators, where fluid dynamics introduce additional nonlinearities and parameter uncertainties. For mobile large-scale manipulators operating in unstructured environments—such as the system illustrated in Fig.~\ref{fig:HHM}—factors like rough terrain, turnover risk, and slippage further complicate control \cite{dadhich2016key}. Among all these issues, communication delay remains one of the most critical challenges in teleoperation. This is especially problematic in field deployments involving large distances and constrained by the latency of wireless or wired communication networks \cite{hasan2025wireless}. These delays can severely degrade both system stability and control performance \cite{lawrence1993stability}.

To address this, there is a vital need for robust control methodologies capable of maintaining stability and achieving accurate tracking of high-level operator commands despite environmental and system uncertainties~\cite{hokayem2006bilateral}. Importantly, such control frameworks do more than ensure operational feasibility—they also enhance the operator's sense of agency by ensuring that robot behavior closely aligns with the operator’s intent. This alignment is a key component of achieving transparency and establishing the SoE in immersive teleoperation for industrial-scale robotic applications. It is important to note, however, that overly aggressive robustness can compromise transparency, creating a mismatch between operator input and perceived robot response, thereby weakening the sense of agency.

\subsection{Impact Resiliency}

Another key control challenge in BHSRMs stems from their high power output and large force interaction range. In unstructured environments, unintended contact can cause severe damage to both the robot and its surroundings. To mitigate this, control frameworks must incorporate collision avoidance and impact-resilient strategies that either prevent contact or minimize its effects. Direct force sensing is ideal for helping this aim but often impractical in large-scale, outdoor systems, prompting reliance on estimation techniques that infer contact from motion deviations, actuator signals, or structural compliance. For instance,~\cite{hejrati2025impact} presents a novel force-sensor-less control method that achieves an 80\% reduction in impact severity for heavy-duty manipulators. However, such estimators introduce latency, as they typically rely on threshold-based detection or filtering, limiting real-time responsiveness of the controller.

To overcome this, improved dynamic modeling via online system identification or vision-based approaches \cite{ko2023vision} may offer a promising path. By enabling predictive adaptation to contact conditions, these models can anticipate interaction forces and trigger earlier responses, enhancing both safety and control performance.

\subsection{Dissimilarity and Scaling}
Teleoperation of BHSRMs is inherently non-intuitive due to significant differences in size, configuration, and mechanical impedance between the human operator and the robot. These dissimilarities constrain the interaction of master–slave robots by primarily relying on end-effector-level commands, as direct joint-to-joint mapping is infeasible. Moreover, due to the mismatch in working space, spatial command scaling becomes necessary to prevent excessive indexing motions during teleoperation. A second, more critical form of scaling pertains to bilateral force-reflecting teleoperation. In these systems, the force output capability of BHSRMs is orders of magnitude greater than human-applied input forces. Consequently, direct one-to-one force reflection can pose serious safety hazards for the operator. As shown in~\cite{daniel1998fundamental}, under perfect force reflection during contact with a rigid environment, the master-side end-effector velocity becomes:

\begin{equation}
    v_m = \left(1 - 2 \dfrac{M_s}{M_m} \right) v,
    \label{vm}
\end{equation}
where \(v_m\) is the master-side end-effector velocity after contact, \(v\) is the contact velocity, and \(M_m\) and \(M_s\) represent the inertia of the master and slave systems, respectively. When \(M_s \gg M_m\), as is typical in BHSRMs, this relationship reveals that unscaled force reflection could transmit dangerously high reaction momentum to the operator.

To address this, high force scaling is required—not only to bring human input into the operational range of the machine but also to mitigate the impact of impedance mismatch between the two systems. While force scaling introduces its own set of stability challenges, it also enables power amplification that by expanding the operator’s functional capability, enables fine control over high-load tasks that would otherwise be physically infeasible—effectively \textit{merging human cognitive intent with robotic mechanical power}.

\section{Designing Immersive Interfaces}
In this section, we examine the design of teleoperation interfaces for BHSRMs with a focus on enhancing operator immersion. Among the various sensory modalities, visual and haptic feedback are the most critical for enabling closed-loop sensorimotor interaction and evoking the SoE. Therefore, our discussion centers on these two modalities, which play a more pivotal role than auditory feedback in the context of large-scale teleoperation.

\subsection{Visual Interface Design}
Visual perception is one of the primary contributors to immersion in teleoperation, particularly when delivered through VR systems~\cite{naceri2019towards}, as the human eye accounts for approximately 70\% of the body's sensory receptors~\cite{cole2021motivated}. Given this dominant role of visual input, enhancing the fidelity and realism of visual feedback is crucial for promoting operator's SoE. Cheng et al.~\cite{cheng2024open} demonstrated that streaming stereoscopic images through a VR headset significantly improved both task completion time and success rate when compared to monocular video, highlighting the role of visual immersion in enhancing teleoperation performance. Similarly, physiological studies such as~\cite{falcone2022pupil} show that pupil dilation correlates with perceived immersion, reinforcing the role of vision in human-in-the-loop systems. Despite this growing body of evidence, only limited efforts—such as~\cite{hejrati2025robust}—have explicitly explored how to leverage visual immersion for BHSRMs.

Visual cues are also essential for evoking the sense of self-location and body ownership—two critical components of SoE. In~\cite{cheng2024open}, an egocentric camera with head-tracking capability was mounted on a humanoid robot’s head at anatomically realistic height to reinforce the illusion of embodiment. However, extending such strategies to BHSRMs is non-trivial due to their dissimilar kinematic structure and scale. A recent study~\cite{hejrati2025robust} showed that mounting a head-tracking camera on the wrist of a BHSRM can still evoke promising levels of SoE, particularly in terms of self-location and ownership, suggesting that embodiment can be induced even in non-anthropomorphic platforms.

Despite these initial insights, open questions remain. To further enhance immersion in BHSRMs teleoperation, one critical research direction is determining optimal camera placement for aligning visual input with the operator's proprioceptive map. This embodiment can be further strengthened by streaming visual input from multiple onboard cameras at the slave site and enabling dynamic viewpoint switching to modulate the field of view as needed, like gaze-based visual interface proposed in \cite{hejrati2025robust}. An alternative or complementary strategy is perceptual scaling of the visual stream through VR with consideration of sense of scale \cite{andrzejczak2021factors}: transformation of the remote robot's image to match human-scale proportions. Such perceptual manipulation may exploit neural mechanisms in the central nervous system to trick the operator into perceiving the large-scale environment as egocentric and familiar, thereby strengthening embodiment and natural control. These ideas open a new space for research at the intersection of robotic vision, cognitive neuroscience, and immersive telepresence.

\subsection{Haptic Interface Design}
\begin{figure}
    \centering
    \includegraphics[width=0.9\linewidth]{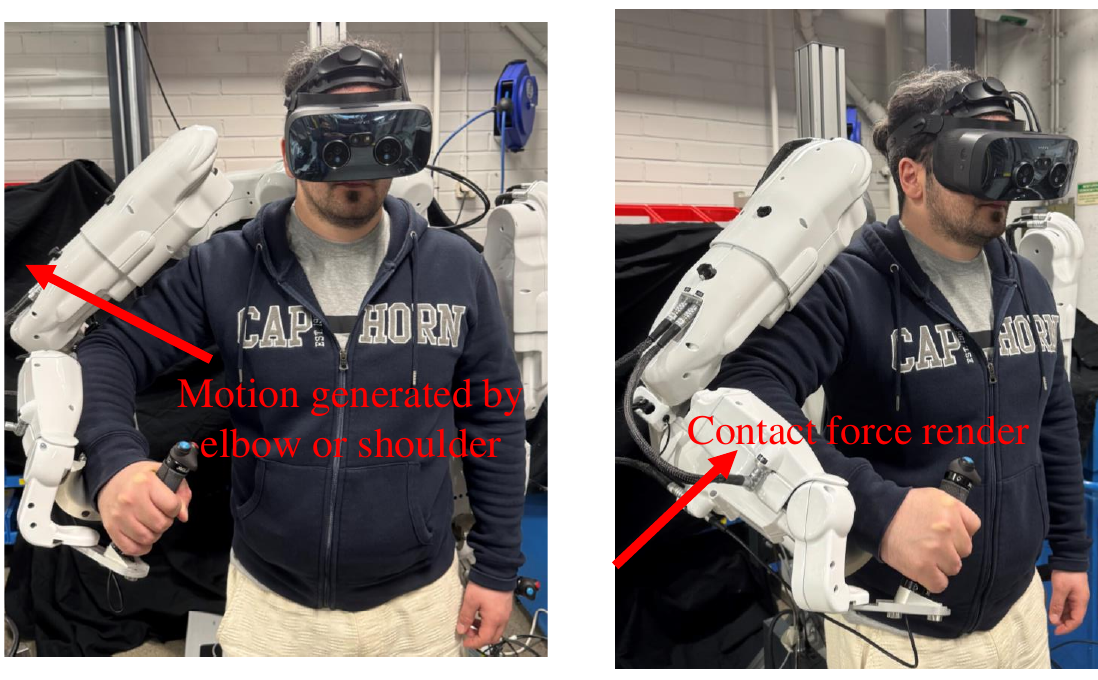}
    \caption{A full-arm haptic exoskeleton used as a master interface in immersive teleoperation. It enables full-arm kinesthetic feedback, supports multimodal data capture, and allows the operator to experience scaled impedance and spatially localized contact forces from the remote environment.}
    \label{exo}
\end{figure}

   
The master interface in a teleoperation system plays a central role in eliciting the Sense of Agency, as it translates human motion into robot commands and establishes a closed sensorimotor loop when coupled with visual feedback. Commonly used master devices include joysticks and haptic displays, which typically engage only the operator’s hand. In contrast, haptic exoskeletons, shown in Fig. \ref{exo}, enable full-arm involvement ~\cite{fang2024airexo}, offering both command input and feedback rendering over the entire kinematic chain \cite{hejrati2025robust}. As illustrated in Fig.~\ref{exo}, haptic exoskeletons integrate the functionalities of both haptic feedback devices and assistive wearable robots, allowing for more immersive and naturalistic interaction. One compelling feature of these systems is their ability to render the scaled impedance of the remote manipulator directly onto the operator's arm~\cite{hejrati2025robust}, establishing a physically grounded connection between the operator and the slave robot. When combined with accurate contact force estimation and localization at the remote site, the exoskeleton can convey spatially relevant force cues to the human body, significantly enhancing immersion and SoE.

Beyond feedback, exoskeletons offer unique advantages for multimodal data collection. Unlike joysticks or point-based haptic devices, they capture full joint trajectories, velocities, and motion strategies used by the operator during task execution. This rich dataset supports high-quality LfD, enabling policy training that reflects not only task intent but also the operator’s biomechanical patterns. In this way, haptic exoskeletons serve as both an expressive control interface and a high-fidelity sensing platform for bridging human skill to BHSRMs.

\section{Human-Centered and Cognitive Challenges}

This section examines human-centered challenges in the immersive teleoperation of BHSRMs. Due to the substantial difference in physical scale and operational context compared to conventional robotic systems, several foundational assumptions in human–robot interaction (HRI) must be reconsidered. Specifically, issues such as operator safety, sensorimotor mismatch arising from incongruencies between visual and proprioceptive cues, and the evaluation of immersion and SoE require new approaches tailored to the unique constraints of BHSRMs.

\subsection{Human Operator Safety}

As shown in Equation~(\ref{vm}), during contact with a rigid environment, the velocity at the master interface may spike proportionally to the inertia ratio between the slave and master systems. In the case of BHSRMs, such as the platforms illustrated in Fig.~\ref{fig:HHM}, this ratio can exceed 50:1 when using a haptic exoskeleton (Fig.~\ref{exo}) as the master device. This discrepancy underscores the importance of appropriate force scaling and control design to ensure operator safety. Moreover, when the operator is physically coupled to the master interface—as in wearable exoskeletons—safety risks extend beyond force reflection to include instability, unintended rapid motions, and potential violations of the anatomical limits of the human arm.

HRI in immersive teleoperation can be broadly divided into physical and cognitive dimensions~\cite{pervez2008safe}. Physical interaction safety encompasses three primary approaches: (i) safety through interaction assessment, (ii) safety through design, and (iii) safety through planning and control ~\cite{pervez2008safe}. The third category was partially addressed in~\cite{hejrati2023physical}, which proposed a model-based control framework augmented with neural networks to maintain robustness while respecting joint limits of the human operator. However, the first two aspects—especially ergonomic design and real-time assessment of interaction risk—require further exploration, both for the slave system and the wearable master interface.

On the cognitive side, Cognitive HRI (cHRI) extends traditional HRI by integrating models of perception, mutual intention, and shared understanding to form a unified cognitive system~\cite{cHRI}. The cHRI emphasizes shared mental models, task-level alignment, and intuitive communication—critical features in immersive teleoperation, where delays, scale mismatches, and remote feedback can distort intent-action congruency. Key design elements include simulation-theoretic models of operator behavior, timing synchronization, social cue integration, and mechanisms for establishing common ground~\cite{cHRI}.

Embedding cHRI principles in teleoperation of BHSRMs would make the sensory information better align with the operator’s internal model, increasing responsiveness, trust, and task fluency, leading to more realistic motor behavior during demonstrations. Therefore, in addition to physical safety and sensory fidelity, cognitive immersion must be treated as a foundational design requirement for scalable, human-centered teleoperation of large-scale robotic manipulators.

\subsection{Sensorimotor Mismatch}
As discussed earlier, motion and force scaling is an essential aspect of teleoperating BHSRMs, helping reduce excessive operator motion and improve task efficiency. However, this scaling introduces not only mathematical and control complexities, but also perceptual conflicts between visual and proprioceptive feedback. Specifically, when the operator’s arm movements are scaled up to match the remote robot's large workspace, the perceived motion of the remote manipulator—fed back via visual cues—may diverge significantly from the felt position of the operator’s own limb.

\begin{figure}[t]
      \centering
      \subfloat[]{\includegraphics[width = 0.45\textwidth]{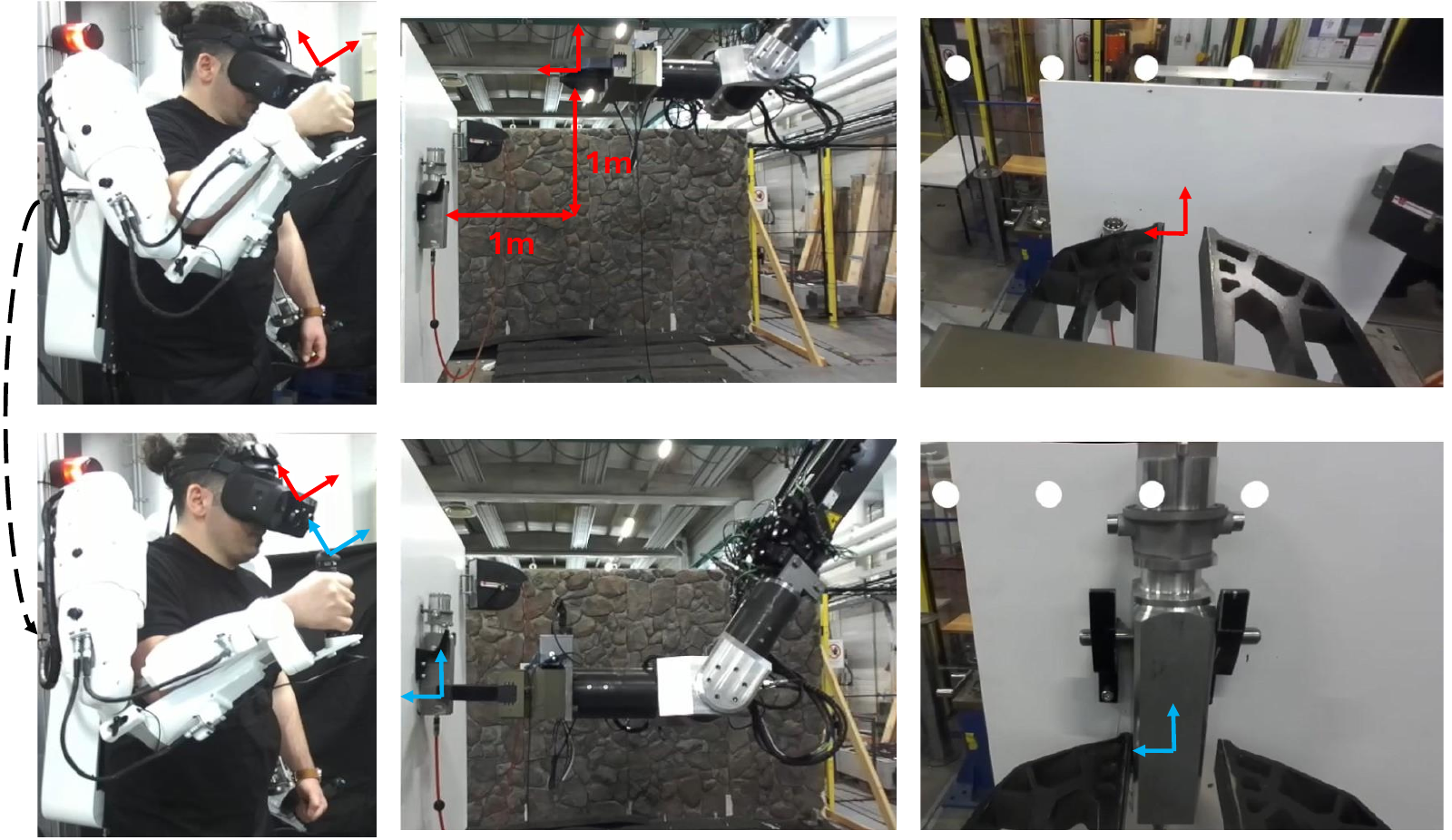}
      \centering
      \label{fig:Teleop}}
      \hfil
      \subfloat[]{\includegraphics[width = 0.35\textwidth]{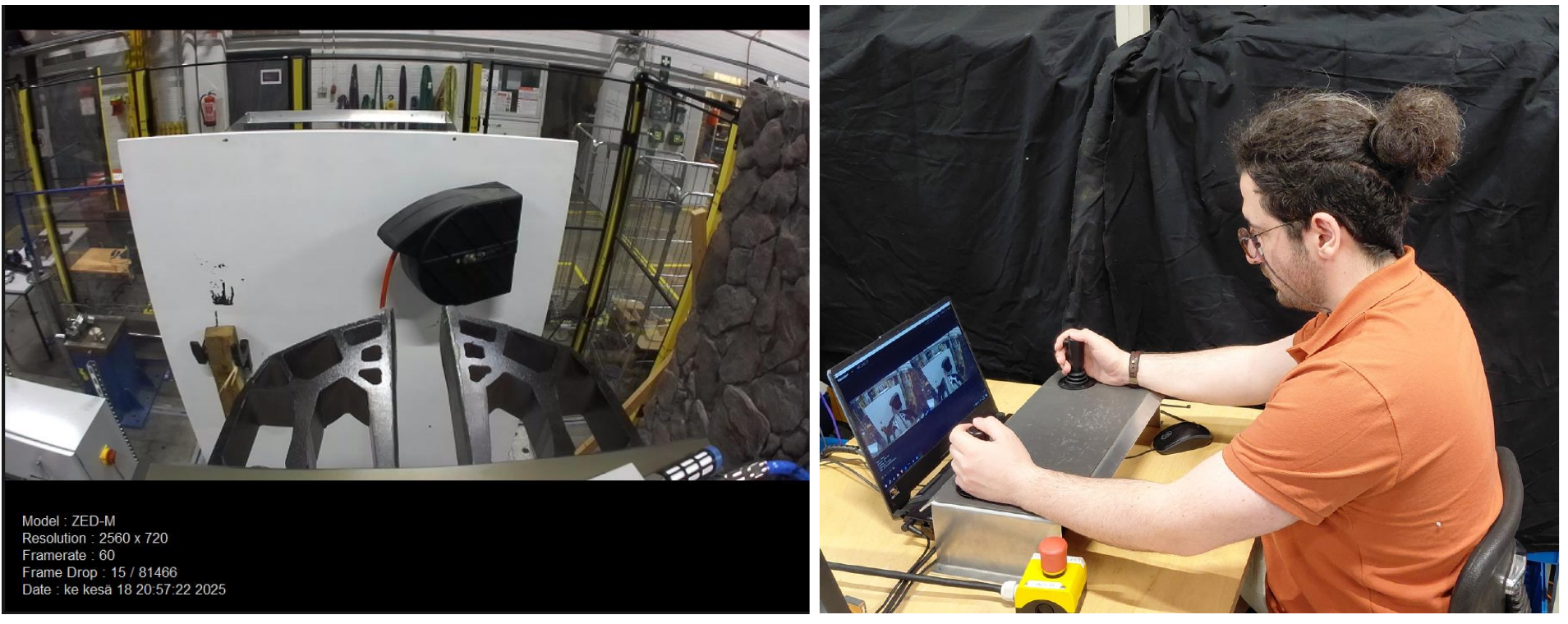}
      \centering
      \label{fig:joysticks}}
      \caption{(a) Immersive teleoperation using a haptic exoskeleton and VR headset with a motion scaling factor of 1:13. The human operator performs minimal arm movement while the remote BHSRM traverses a large spatial range. (b) Teleoperation using a conventional joystick interface and 2D monitor. This condition offers lower immersion and limited proprioceptive engagement.}
      \label{HULE confg}
   \end{figure}

To investigate this phenomenon, an immersive teleoperation experiment was conducted as a case study with a motion scaling factor of 1:13, using a haptic exoskeleton and a VR headset with head tracking. As illustrated in Fig.~\ref{fig:Teleop}, the upper row shows the initial pose, while the lower row displays the target pose after a successful grasp. Despite only a small displacement of the operator’s arm (from red to blue marker), the remote robot traversed nearly one meter. The operator reported an initial sense of disorientation: minor movements of their own limb produced disproportionately large movements in the remote view. This conflict between visual feedback and proprioceptive sensation represents a sensory mismatch. Such mismatch is particularly pronounced in high-immersion settings where the entire arm is involved, as opposed to low-immersion setups like joystick interfaces, shown in Fig.~\ref{fig:joysticks}, where control is limited to the fingers and does not engage the operator’s full body schema. These initial results suggest that immersive interfaces—while enhancing SoE—also increase the likelihood and salience of sensorimotor mismatch, especially under aggressive scaling. Further investigation is needed to characterize the extent and impact of this phenomenon in BHSRMs teleoperation.

\section{Future Directions and Research Opportunities}
As immersive teleoperation of BHSRMs continues to evolve \cite{liu2023multi, lee2022challenges}, several promising directions warrant further investigation to overcome current limitations. The following areas highlight potential pathways for enhancing control robustness, operator embodiment, and cognitive integration in large-scale robotic systems.

\subsection{Scalable Control for High-DoF Teleoperation}

One of the most promising control methodologies for BHSRMs is the Virtual Decomposition Control (VDC) scheme~\cite{zhu2010virtual}. By decomposing complex, high-degree-of-freedom systems into manageable subsystems using a baseline Newton–Euler model, VDC significantly reduces computational complexity while maintaining modularity. This makes it particularly suitable for real-time control in immersive teleoperation scenarios. The effectiveness of VDC has been validated through implementation on real-world BHSRMs~\cite{koivumaki2015stability, hejrati2025orchestrated, koivumaki2019energy} with established sub-centimeter tracking accuracy. More recently, it has been successfully deployed on a 6-DoF industrial manipulator controlled via a 7-DoF haptic exoskeleton interface, enabling immersive teleoperation with high fidelity~\cite{hejrati2025robust}. These promising results suggest that VDC provides a computationally efficient, scalable and robust foundation for future telepresence architectures in large-scale systems.

In addition, incorporating elements of shared autonomy—such as those proposed in~\cite{luo2022human}—can enhance performance under challenging control conditions by providing assistive behaviors that support the operator without overriding manual input. This approach preserves the operator’s sense of agency while improving task success and operational safety.

\subsection{Mitigating Sensorimotor Mismatch}
As immersive interfaces advance, addressing the perceptual conflicts caused by aggressive motion scaling and full-body interaction becomes increasingly critical in BHSRMs teleoperation. Sensorimotor mismatches—such as discrepancies between proprioceptive feedback and visual cues—can impose cognitive load and impair task fluency. Future work should aim to quantify these effects by conducting more user studies, model user adaptation, and develop real-time strategies for sensory alignment. One promising approach lies in the optimal information and effort model proposed by Cheng et al.~\cite{cheng2025human}, which describes how human motor behavior adapts to unimodal and multimodal sensory feedback by balancing sensory informativeness against neuromuscular effort. Applying this framework to teleoperation may offer a principled means to tune control interfaces or feedback modalities, thereby helping the central nervous system resolve sensory discrepancies more efficiently during remote manipulation. Other potential approaches include adaptive visual remapping based on user behavior and personalized scaling profiles that dynamically adjust the mapping between master and slave systems. 

\subsection{Evaluating and Quantifying SoE at Scale}

Another critical area requiring attention is the development of rigorous and scalable methods for evaluation of SoE, particularly in the context of BHSRMs. While current assessment techniques include both implicit measures—such as task completion time and physiological indicators like pupil dilation~\cite{falcone2022pupil}—and explicit measures using questionnaires, these tools have largely been developed for human-scale or VR-based interfaces. Critically, there is no validated questionnaire designed specifically for the unique context of BHSRMs. Their extreme scale, dissimilar morphology, and dynamic complexity present sensory and cognitive conditions not addressed by existing SoE frameworks. Although~\cite{hejrati2025robust} employed a custom questionnaire in an initial study, this remains a preliminary effort. A robust evaluation framework tailored to BHSRMs must account for proprioceptive scaling, sensorimotor mismatch, non-anthropomorphic embodiment, and operator trust under high workload and delayed feedback.

Developing such tools is essential not only for interface design, but also for benchmarking immersion quality across platforms, tracking user adaptation, and informing learning-from-demonstration frameworks that depend on high-quality human input.

\section{Conclusion}
This paper explored the unique challenges of immersive teleoperation in BHSRMs, highlighting the critical role of control design, interface immersion, and cognitive alignment. Through conceptual analysis and early-stage experiments, we demonstrated how high degrees of immersion—achieved via haptic exoskeletons and visual feedback—can introduce sensorimotor and safety challenges. We emphasized the importance of human-centered design in achieving realistic, intuitive, and scalable teleoperation and outlined key directions for future work, including SoE evaluation tools and adaptive feedback strategies to mitigate sensory motor mismatch. These insights aim to lay the groundwork for the next generation of large-scale, learning-enabled telepresence systems.

\bibliographystyle{IEEEtran}
\bibliography{mybib}

\end{document}